\documentclass[11pt,a4paper]{article}
\usepackage[hyperref]{acl2021}
\usepackage{times}
\usepackage{latexsym}

\usepackage{microtype}
\usepackage{latexsym}
\usepackage{mathtools}
\usepackage{bbm}
\usepackage{dutchcal}
\usepackage{graphicx}
\usepackage{amsmath}
\usepackage{subcaption}

\usepackage{microtype}
\usepackage{multirow}
\usepackage{pifont}
\usepackage[T1]{fontenc}
\usepackage{lineno}
\usepackage{ctable}
\usepackage[normalem]{ulem}
\usepackage[bottom]{footmisc}
\usepackage{float}

\makeatletter
\newcommand{\xdashrightarrow}[2][]{\ext@arrow 0359\rightarrowfill@@{#1}{#2}}
\newcommand{\xdashleftarrow}[2][]{\ext@arrow 3095\leftarrowfill@@{#1}{#2}}
\newcommand{\xdashleftrightarrow}[2][]{\ext@arrow 3359\leftrightarrowfill@@{#1}{#2}}
\def\rightarrowfill@@{\arrowfill@@\relax\relbar\rightarrow}
\def\leftarrowfill@@{\arrowfill@@\leftarrow\relbar\relax}
\def\leftrightarrowfill@@{\arrowfill@@\leftarrow\relbar\rightarrow}
\def\arrowfill@@#1#2#3#4{%
  $\m@th\thickmuskip0mu\medmuskip\thickmuskip\thinmuskip\thickmuskip
   \relax#4#1
   \xleaders\hbox{$#4#2$}\hfill
   #3$%
}
\makeatother

\usepackage{amsthm,amsmath,amsfonts,bm,xspace}
\usepackage{upgreek}
\usepackage{color}

\newcommand{\comment}[1]{}

\def\eqref#1{(\ref{#1})}

\def\1{\bm{1}}

\def\vtheta{{\bm{\theta}}}

\def\vf{{\bm{f}}}

\DeclareMathAlphabet{\mathsfit}{\encodingdefault}{\sfdefault}{m}{sl}
\SetMathAlphabet{\mathsfit}{bold}{\encodingdefault}{\sfdefault}{bx}{n}
\newcommand{\tens}[1]{\bm{\mathsfit{#1}}}

\def\tW{{\tens{W}}}

\usepackage{microtype}

\aclfinalcopy 
\def\aclpaperid{4178} 

\title{Neural-Symbolic Commonsense Reasoner with Relation Predictors}

\author{Farhad Moghimifar$^1$ \and \bf{Lizhen Qu}$^2$ \and \bf{Yue Zhuo}$^3$ \\
         \bf{Gholamreza Haffari}$^2$ \and \bf{Mahsa Baktashmotlagh}$^1$\\
         $^1$The School of ITEE, The University of Queensland, Australia\\
         $^2$Faculty of Information Technology, Monash University, Australia\\
         $^3$School of CSE, The University of New South Wales,  Australia\\
         \tt \{f.moghimifar,m.baktashmotlagh\}@uq.edu.au\\
         \tt firstname.lastname@monash.edu, 
         \tt terry.zhuo@unsw.edu.au
         }

\date{}

\begin{document}
\maketitle

\begin{abstract}
Commonsense reasoning aims to incorporate sets of commonsense facts, retrieved from Commonsense Knowledge Graphs~(CKG), to draw conclusion about ordinary situations. 
The dynamic nature of commonsense knowledge postulates models capable of performing multi-hop reasoning over new situations. This feature also results in having large-scale sparse Knowledge Graphs, where such reasoning process is needed to predict relations between new events. However, existing approaches in this area are limited by considering CKGs as a limited set of facts, thus rendering them unfit for reasoning over new unseen situations and events. In this paper, we present a neural-symbolic reasoner, which is capable of reasoning over large-scale dynamic CKGs. The logic rules for reasoning over CKGs are learned during training by our model. In addition to providing interpretable explanation, the learned logic rules help to generalise prediction to newly introduced events. Experimental results on the task of link prediction on CKGs prove the effectiveness of our model by outperforming the state-of-the-art models.

\end{abstract}

\section{Introduction}
Commonsense reasoning refers to the ability of capitalising on commonly used knowledge by most people, and making decisions accordingly~\citep{sap2020introductory}. This process usually involves combining multiple commonsense facts and beliefs to draw a conclusion or judgement~\citep{lin2019kagnet}. While human trivially performs such reasoning, current Artificial Intelligence models fail, mostly due to challenges of acquiring relevant knowledge and forming logical connections between them. 

Recent attempts in empowering machines with the capability of commonsense reasoning are mostly centred around large-scale Commonsense Knowledge Graphs~(CKG), such as ATOMIC and ConceptNet~\citep{sap2019atomic, speer2017conceptnet}. Unlike conventional Knowledge Graphs~(KG), CKGs usually contains facts about arbitrary phrases. For instance, ``PersonX thanks PersonY" is connected to `` To express gratitude" via the link `` because X wanted". This non-canonicalised free-form text representation has resulted in having conceptually related nodes with different representation, which forms \emph{large sparse} CKGs~\citep{malaviya2020commonsense}. Therefore, established reasoning models on conventional KGs perform poorly on CGKs~\citep{yang2014embedding, sun2018rotate, dettmers2018convolutional, minervini2020learning}. In addition, the nature of commonsense reasoning encourages dynamic CKGs, where new sets of facts and phrases are introduced frequently. Most existing models in this realm are based on a static set of facts and phrases, which results in poor generalisation~\citep{malaviya2020commonsense, shang2019end}. Nevertheless, the inference process in existing approaches is like a \emph{black box}, where internal behaviour of the model is hardly interpretable.

To overcome these limitations, we propose a neural-symbolic reasoning model based on backward-chaining. While traditional theorem proving algorithms~\citep{bratko2001prolog} work based on a set of predefined rules and unification over discrete symbols, we leverage a continuous relaxation of weak unification and a rule learner module. The weak unification over continuous embedding representation helps to address the challenges of unseen sparsity of CKGs. The rule learner module, in addition to providing interpretability, is used to generalise prediction to unseen data points to mitigate the problem of \emph{large-scale dynamic} CKGs. The experiments on the task of link prediction confirm the superiority of our model, by a margin of up to 22 points, over the state-of-the-art models.

\begin{figure*}[ht]
    \centering
    \includegraphics[width = 0.9\textwidth]{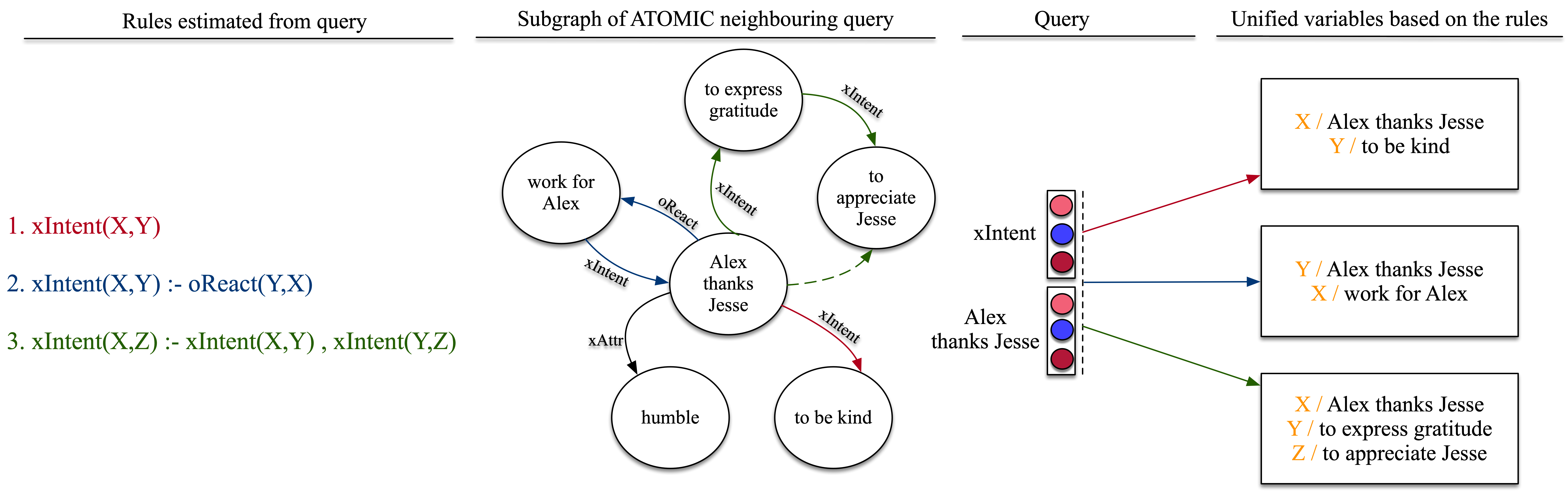}
    \caption{A visual representation of rules and new relations estimated by our model for a sample query, \emph{xIntent(Alex thanks Jesse, ?}. Based on the subject of the query, a subgraph of ATOMIC is retrieved for the reasoning process (middle). Sets of rules estimated from relation of the query is generated using our proposed rule creation module~(left). Based on our reasoning model the answers to query are predicted by unification module~(left).}
    \label{fig:model}
\end{figure*}

\section{Related Works}
Recent approaches in knowledge base completion task have mostly relied on a graph and entity-relation embedding methods~\citep{yang2014embedding, dettmers2018convolutional}. In these approaches, entities and relations are embedded in a complex space, and using a scoring function plausibility of a triple is estimated~\citep{bordes2013translating,trouillon2016complex,sun2018rotate}. In addition to node embedding, graph embedding methods have also been used to capture the structural complexity of knowledge bases~\citep{schlichtkrull2018modeling,shang2019end}. 
Language generative models also have been applied on knowledge bases in order to use the rich information of pre-trained models to address CKG completion task~\citep{bosselut2019comet, moghimifar2020cosmo}. \citet{malaviya2020commonsense} proposed a method based on using language models and graph networks to solve the problem of the sparsity of CKGs, by taking structural and contextual characteristics of CKGs into account. However, the aforementioned models are highly dependant on training on a set of static entities, and fail to perform when new triples are presented.
\vspace{-1ex}

\section{Our Approach}
\vspace{-2ex}
A CKG $\mathcal{G}=(\mathcal{N},\mathcal{E})$, where $\mathcal{N}$ is the set of nodes and $\mathcal{E}$ is the set of edges in $\mathcal{G}$, consists of triples in form of $r(h,t)$, where $h,t \in \mathcal{N}$ are referred to as the head and the tail of the triple, and $r \in \mathcal{E}$ denotes their relation. The goal of the CKG completion task is to estimate probable 
$t$ given a query $q=r(h, ?)$. As the target node may not pose a direct link to $h$ via $r$, this task requires a model capable of multi-step reasoning.

\begin{table*}[ht]
    \centering
    \resizebox{0.9\textwidth}{!}{
    \begin{tabular}{c c c c c c c c }
        \specialrule{.1em}{.1em}{.1em}
        Dataset & \#Nodes & \#Edges  & Avg. In-degree & Density & Unseen Nodes & Unseen Edges & \#Relations\\
        \hline
         ATOMIC & 382823 & 785952  & 2.25 & 1.6e-5 & 38.36\% & 27.91\% & 9\\
         ConceptNet-100k & 80994 & 102400  & 1.25 & 9.0e-6 & 11\% & 8\% & 34 \\
         \specialrule{.1em}{.1em}{.1em}
    \end{tabular}}
    \caption{Statistics on ATOMIC and ConceptNet-100k. Unseen Nodes is the ratio of the nodes in test set that are not in train set to all of the nodes in test set. Unseen edges is the ratio of edges where either the head or tail nodes are not in train set to the number of all edges in test set.}
    \label{tab:ckg_stat}
    \vspace{-1ex}
\end{table*}

Given a query $r_q(h_q, ?)$, we try to identify an implication rule and apply it to prove $r_q(h_q, t)$ for a target entity or event $t$. A rule $\mathcal{R}$ takes the form of $r_q (X, Z) \mathop{:\!-} r_0(X, Y_0), ... , r_k(Y_{k-1}, Z)$, where capitalised letters denote variables, $r_q (X, Z)$ is the rule head, and the rule body is a conjunction of atoms. We apply such a rule by unifying atoms with triples in the given CKG to obtain $r_q (h_q, t_k) \mathop{:\!-} r_0(h_0, t_0) , r_1(t_0, t_1) , ... , r_k(t_{k-1}, t_k)$, which entails $r_q (h_q, t_k)$. 
Since semantically equivalent/similar events or entities in a CKG often have different surface forms, we consider weak unification of an atom with a triple
Instead of only considering exact match of two atoms, a weak unification operator~\cite{sessa2002weakUnify} unifies two different symbols by measuring the similarity of their representations.

Given a query, we do not know the target rule in advance. As shown in the example in Fig. \ref{fig:model}, we successively create a new rule by appending the body of the previous rule with an atom in the form of $r(t_{k-1},X)$. Whenever such a new atom is added, we query the CKG to find triples as candidates of unification. This step enables reasoning on large scale KBs. In contrast, the prior work~\citep{minervini2020learning,ren2020beta} require comparison with each node in a CKG. After applying the weak unification operator to each of the triples, we find top$k$ most similar nodes and use each of the entity/event in the place of $X$ to create a new atom for a new rule. The process is repeated until the maximal rule length is reached.

The above mentioned reasoning process is delivered by a a neural-symbolic reasoner. It consists of a query module, a weak unification operator, and a rule creation module. 
\vspace{-2ex}
\paragraph{Query} Given a rule with a rightmost atom $r_{k-1}(t_{k-1},X)$ in the rule body, we send the representation of $t_{k-1}$ as query to the given CKG to retrieve unification candidates. A node in a CKG is a word sequence. To support comparison of nodes w.r.t. their semantic similarities, we encode queries and nodes in a CKG with a pre-trained BERT~\citep{devlin2019bert} into embeddings. To this end, a node is converted into $[CLS] + node + [SEP]$, and fed into the model, and we use the representation of $[CLS]$ token from the last layer of BERT as representation of node $node$. We apply FAISS\footnote{https://github.com/facebookresearch/faiss}~\citep{johnson2019billion} to index embeddings of an CKG, because it supports fast retrieval of $k$ nearest neighbours of a dense vector. For each node $v$ in the top$k$ list, we collect a set of triples $\mathcal{C}(v)$, which are all triples having $v$ as the head in the CKG. As a result, we have $k$ such sets and form a candidate set $\mathcal{C}$ by taking the union of them.

\paragraph{Weak Unification} From a candidate set $\mathcal{C}$ we identify top$k$ most relevant triples to unify $r_{k-1}(t_{k-1},X)$. First, we formulate a set of hypotheses $\mathcal{H}$ by replacing $X$ with possible tails. In practice, we use all tails of the triples in $C$. Furthermore, we construct a bipartite graph between $\mathcal{C}$ and $\mathcal{H}$, in which an edge denotes the unification between a triple from $\mathcal{C}$ and another from $\mathcal{H}$. We measure unification scores by using cosine similarity and obtain an similarity matrix $\mathbf{U} \in \mathbb{R}^{|\mathcal{C}| \times |\mathcal{H}|}$. The final unification score of candidate triple $i$ is computed by $\max_j\mathbf{U}_{ij}$. We keep only top$k$ highest scored candidate triples. 
\vspace{-1ex}
\paragraph{Rule creation} Given the top$k$ highest scored candidate triples and a rule $\mathcal{R}_k$ with a rightmost atom $r_{k-1}(t_{k-1},X)$, we create a new rule based on $\mathcal{R}_k$ for each triple $k$ by substituting it for $r(t_{k-1},X)$ and append another atom $r_k(t_k,X)$. The relation $r_k$ is estimated by a relation predictor $\vf_{\vtheta}(r_{k-1}, k)$, where both $r_{k-1}$ and the current step $k$ are mapped to the corresponding embeddings.
\vspace{-1ex}
\begin{equation}
    P_{\vtheta_\vf}(r_k|r_{k-1},k) = \sigma(\vf_{\vtheta}([r_{k-1};k]).\tW + b)
\end{equation}

where $\vtheta_{\vf}:=\{\tW,b\}$ contains the Rule creation module's parameters, and $\sigma$ is the sigmoid function. The relation predictor aims to generalise relation co-occurrence patterns in rules. We implement it by using a neural networks with two blocks of hidden layers, followed by a softmax layer. Each block is composed of a linear layer and a ReLU layer.

Given a query $r_q(h_q, ?)$, we initialise the first rule as $r_q(h_q, X)$. After reaching the pre-defined maximal rule length, we consider the score of a rule after unification as the lowest unification score associated with the rule, following~\cite{sessa2002weakUnify}. We rank all rules by their scores and select the tails in the rule heads of the top$k$ highest scored rules as the results. 

Another benefit of our reasoner is that humans can easily collect evidences to interpret reasoning results. The model can yield the rules and unified triples in a human-friendly format, which are generated at each step. In contrast, prior work~\cite{malaviya2020commonsense} on commonsense reasoners produces only hard-to-understand distributed representations in intermediate steps.  
\vspace{-1ex}
\paragraph{Training} 
We convert all the triples in $\mathcal{G}$ into a set of queries~($\mathcal{Q} = \{r_1(h_1,?), r_2(h_2,?),\dots, r_n(h_n,?),\} $), where each query of $r_i(h_i,?) (i<n)$ is associated with a set of gold answers $\mathcal{T_i} = \{q_{i_1}, q_{i_2},\dots, q_{i_m}\}$. The goal of training our model is to learn the embedding representations by minimising a cross-entropy loss function~($\mathcal{L}_\theta$) on final scores associated with each estimated predictions and the set of gold answer:  
\vspace{-1ex}
\begin{align}
    \mathcal{L}_{\vtheta} = -& \sum_{q_{p_k}\in \mathcal{T}} \log(Pr(q_{p_k}|\mathcal{G}; \vtheta)) \label{eq:obj} \\ 
                      -& \sum_{q_{p_k} \not\in \mathcal{T}} \log(1- Pr(Pr(q_{p_k}|\mathcal{G}; \vtheta))) \nonumber
\end{align}

where $\vtheta$ denote all the parameters of our model. The relation predication module of our model is also trained by minimising loss in equation \ref{eq:obj}, where the relation embeddings are decoded by alignment of the associated embedding and nearest predicate representation.

\section{Experiments}
\begin{table*}[ht]
    \centering
        \resizebox{0.9\linewidth}{!}{
    \begin{tabular}{l c c c c c c c c}
        \specialrule{.1em}{.1em}{.1em} 
         & \multicolumn{4}{c}{\textbf{ConceptNet-100k}} & \multicolumn{4}{c}{\textbf{ATOMIC}} \\
         \cline{2-9}
         \textbf{Model} & MRR & HITS@1 & HITS@3 & HITS@10 & MRR & HITS@1 & HITS@3 & HITS@10 \\
         \hline
         DistMult & 8.68 & 5.38 & 9.33 & 15.23 & 11.49 & 9.16 & 11.83 & 16.3 \\
         ComplEx & 10.33 & 6.51 & 11.24 & 17.31 & 12.96 & 10.65 & 13.9 & 17.08 \\
         ConvE & 16.55 & 10.19 & 18.79 & 28.08 & 9.04 & 7.05 & 9.42 & 12.74 \\
         RotatE & 19.89 & 14.45 & 25.32 & 37.56 & 10.61 & 8.56 & 10.76 & 14.98 \\
         Malaviya et al. & 43.60 & 39.33 & 49.41 & 66.58 & 23.43 & 20.54 & 24.1 & 27.43 \\
         \hline
         \textbf{Ours} & \textbf{65.72} & \textbf{57.49} & \textbf{61.7} & \textbf{71.46} & \textbf{46.41} & \textbf{43.31} & \textbf{45.94} & \textbf{47.24} \\
         \specialrule{.1em}{.1em}{.1em}
    \end{tabular}}
    \caption{Results on CKG completion task, on ConceptNet-100K and ATOMIC.}
    \label{tab:ckg_completion}
    \vspace{-1ex}
\end{table*}

To evaluate the performance of our model~\footnote{Code available at \url{https://github.com/farhadmfar/commonsense\_reasoner}} in the task of CKG completion, in this section, we report the results of our model in comparison with the baselines.

\noindent \textbf{Evaluation Metrics:} Following previous works on Knowledge Base completion~\citep{dettmers2018convolutional,malaviya2020commonsense}, we report the results of HITS and Mean Reciprocal Rank. Similar to \citet{dettmers2018convolutional}, when computing the scores for a gold target entity, we filter out all remaining valid entities. Furthermore, for each triple \emph{$(h,r,t)$}, the score is the average of scores measured from \emph{$(h,r,?)$} and \emph{$(t,r^{-1},?)$}. 

\noindent \textbf{Baselines} For comparison, we report the performance of state-of-the-art models in CKG and KB completion. We compare our model to DistMult~\citep{yang2014embedding}, ComplEx~\citep{trouillon2016complex}, ConvE~\citep{dettmers2018convolutional}, RotatE~\citep{sun2018rotate}, and Malaviya~\citep{malaviya2020commonsense}. The first four models are high performance models in conventional KB completion, whereas the latter is proposed for CKG completion.
\vspace{-1ex}
\subsection{Datasets}
\noindent \textbf{ATOMIC}~\footnote{https://homes.cs.washington.edu/~msap/atomic/} is a CKG consisting of commonsense facts in form of triples, based on \emph{if-then} relations~\citep{sap2019atomic}. This dataset consists of more 877K facts, and more than 300K entities. 

\noindent \textbf{ConceptNet-100K}~\footnote{https://ttic.uchicago.edu/~kgimpel/commonsense.html} is a subset of ConceptNet 5~\citep{speer2017conceptnet}, containing Open Mind Common Sense~(OMCS) entries, introduced by~\cite{li2016commonsense}. This dataset contains general commonsense facts in form of triples.

In order to evaluate the performance of the models in dynamic CKG completion, we choose a subset of the test set of ATOMIC and ConceptNet-100K, where for any \emph{$(h,r,t)$} either \emph{$h$} or \emph{$t$} is not seen by the model in the train set. Statistics on ATOMIC and ConceptNet-100k are provided in table~\ref{tab:ckg_stat}.
\vspace{-1ex}
\subsection{Results}

Table \ref{tab:ckg_completion} summarises the results of the conducted experiment on ConceptNet-100K and ATOMIC. On ConceptNet-100K our proposed model outperforms the baselines by up to 22 points on MRR. The gap between our model and the second best model decrease as we move from HITS@1 to HITS@10. This suggested that on contrary to the baselines our model performs better in estimating the probability of query with higher accuracy. On ATOMIC our model achieves a MRR of 46.41, which is 13 points higher than the second best model. As it can be seen from table \ref{tab:ckg_completion}, comparison of performance of different models on ConceptNet-100K and ATOMIC shows a noticeable drop in performance for models which rely on structural information of CKGs. This observation suggests that larger and sparser~(lowest density) CKG are more challenging to reason over.

Table \ref{tab:rules} provides examples of generated rules by our model on ATOMIC and ConceptNet-100k. On ATOMIC, the first rule is based on transition, and the second and third rules are inverse rules. Similarly, on ConceptNet-100K the first and third rules are transitive, and the second rule is a compositional rule. All provided rules are diverse and meaningful, and can be used for explaining the inference process of our model. Some examples of interpretability of our model can be found in Appendix 1.

\begin{table}[h]
    \centering
    \resizebox{\linewidth}{!}{
    \begin{tabular}{c}
        \specialrule{.1em}{.1em}{.1em}
         \textbf{ATOMIC} \\
         \specialrule{.1em}{.1em}{.1em}
          {\fontfamily{qcr}\selectfont
            xIntent(X,Y):-xIntent(X,Z),xIntent(Z,Y)}\\
            {\fontfamily{qcr}\selectfont xNeed(X,Y):-xReact(Y,X)}\\
            {\fontfamily{qcr}\selectfont xIntent(X,Y):-oWant(Y,X)}\\
            \specialrule{.1em}{.1em}{.1em}
            \textbf{ConceptNet-100K} \\
            \specialrule{.1em}{.1em}{.1em}
            {\fontfamily{qcr}\selectfont
            causes(X,Y):-causes(X,Z),causes(Z,Y)}\\
            {\fontfamily{qcr}\selectfont
            isa(X,Y):-partof(X,Z),isa(Z,Y)}\\
            {\fontfamily{qcr}\selectfont
            relatedto(X,Y):-relatedto(X,Z),relatedto(Z,Y)}\\
            \specialrule{.1em}{.1em}{.1em}
    \end{tabular}}
    \caption{Examples of rules learned by our proposed relation prediction module.}
    \label{tab:rules}
    \vspace{-1ex}
\end{table}

\section{Conclusion}
\vspace{-1ex}
In this work, we propose a neural-symbolic reasoning model over Commonsense Knowledge Graphs~(CKGs). Our proposed model leverages a relation prediction module, which provide capability of multi-step reasoning. This ability, alongside weak unification, helps generalising our model to large-scale unseen data. We showed that our model yields state-of-the-art results when applied to large-scale sparse CKGs, and the inference step is interpretable.

\bibliographystyle{acl_natbib}
\bibliography{anthology,acl2021}

\appendix
\newpage
\section{Examples of Interpretability of our Model}

One of the advantages of our model, in comparison to other models in this area, is the interpretability of the inference. Based on learned rules during training, the model takes a path to reach an answer node in CKGs. For instance, consider a query of \emph{xIntent(Alex drives Jesse there, ?)}. Based on first rule from Table 3, {\fontfamily{qcr}\selectfont
X}
is unified by \emph{Alex drives Jesse there}, and {\fontfamily{qcr}\selectfont
Z} 
is unified by \emph{Alex helps Jesse}~(from triples of ATOMIC). Then, the query is updated to \emph{xIntent(Alex helps Jesse, ?)} and  {\fontfamily{qcr}\selectfont
Y}
is unified by \emph{to be of assistance}~(from triples of ATOMIC), hence the answer to query. 
The path generated by this example is \emph{Alex drives Jesse there} $\xrightarrow[]{ \text{xIntent}}$ \emph{Alex helps Jesse} $\xrightarrow[]{ \text{xIntent}}$ \emph{to be of assistance}. Therefore, two nodes are connected via a new link:
\emph{Alex drives Jesse there} $\xdashrightarrow[] {\text{xIntent}}$ \emph{to be of assistance}.

Consider the following query from ConceptNet-100K, $HasProperty(novel, ?)$. Based on the relation of the query, our rule creator module can estimate the following rule:

\begin{table}[h]
    \centering
    \resizebox{\linewidth}{!}{
    \begin{tabular}{l}
    {\fontfamily{qcr}\selectfont
            HasProperty(X,Y):-IsA(X,Z),HasProperty(Z,Y)}
    \end{tabular}}
    \label{tab:my_label}
\end{table}

\vspace{-2ex}
According to this rule, 
{\fontfamily{qcr}\selectfont
X}
is unified by \emph{novel}, and {\fontfamily{qcr}\selectfont
Z} 
is unified by \emph{book}~(from triples of ConceptNet-100K). Then, the query is updated to \emph{HasProperty(book, ?)} and  {\fontfamily{qcr}\selectfont
Y}
is unified by \emph{expensive}~(from triples of ConceptNet-100K), resulting the answer to the query, by generating the following path:
\emph{novel} $\xrightarrow[]{ \text{IsA}}$ \emph{book} $\xrightarrow[]{ \text{HasProperty}}$ \emph{expensive}, hence \emph{novel} $\xdashrightarrow[]{ \text{HasProperty}}$ \emph{expensive}.

\section{Hyperparameter Selection}

To train our model, each triple in form $r(h,t)$ in train set was converted to $r^{-1}(t,h)$, to account for reverse relations as well. We have used the embedding size of 1024 for both node and relation embedding layer. To embed the nodes in CKGs, we have fine-tuned uncased BERT-Large~\citep{devlin2019bert} for the objective of masked language model. For this purpose, a node is converted into $[CLS] + n_i + [SEP]$ and fed into BERT. The representation of the token $[CLS]$ from last year of BERT is then used as node $n_i$ embedded representation. We used the maximum sequence of 128, and batch size of 64. Our relation predication module consists of two Linear layer. For all non-linearities in our model we have used ReLU. For optimisation purpose, SGD has been used, with staring learning rate of $10e-4$, and decay rate of 0.9, if the loss of development set does not decrease after each epoch. We set the maximum depth of three for reasoning process. We have trained the model for 100 epochs.

Followed by \citet{malaviya2020commonsense}, we have trained all the baselines for 200 epochs. During training the models were evaluated on development set, every 10 and 30 epochs, for ConceptNet-100K and ATOMIC, respectively. The checkpoint with the highest MRR was then selected for testing.

\end{document}